%% file: paper.tex
\documentclass[]{style}

\input{preamble}

\usepackage[utf8]{inputenc} 
\usepackage[T1]{fontenc}    
\usepackage{graphicx}
\usepackage{amsmath}
\usepackage{amssymb}
\usepackage{url}            
\usepackage{booktabs}       
\usepackage{amsfonts}       
\usepackage{nicefrac}       
\usepackage{microtype}      
\usepackage{xcolor}         
\usepackage{multirow}
\usepackage{comment}
\usepackage{colortbl}
\usepackage{tocloft}
\usepackage{algorithm}
\usepackage{algorithmicx}
\usepackage{algpseudocode}
\usepackage{wrapfig}
\usepackage{tabularx}
\usepackage{textcomp}
\usepackage{stfloats}
\usepackage{verbatim}
\usepackage{titlesec}
\usepackage{adjustbox}
\usepackage{pifont}
\usepackage{tikz}
\usepackage{color}
\usepackage{subcaption}
\usepackage{soul}

\usepackage{makecell}
\newcolumntype{Y}{>{\centering\arraybackslash}X}
\usepackage{multirow}
\usepackage[table]{xcolor} 
\definecolor{bestcolor}{HTML}{A9D18E} 
\definecolor{sbestcolor}{HTML}{E2EFDA}
\newcommand{\best}{\cellcolor{bestcolor}}
\newcommand{\sbest}{\cellcolor{sbestcolor}}
\newcolumntype{C}{>{\centering\arraybackslash}m{1.3cm}}
\newcolumntype{Z}{>{\raggedright\arraybackslash\fontsize{6.4pt}{7.2pt}\selectfont}X}

\RequirePackage{xspace}
\makeatletter
\DeclareRobustCommand\onedot{\futurelet\@let@token\@onedot}
\def\@onedot{\ifx\@let@token.\else.\null\fi\xspace}

\makeatother

\definecolor{lightblue}{rgb}{0.66, 0.85, 0.95}
\hypersetup{
    colorlinks=true,
    linkcolor=red,
    citecolor=blue,
}
\definecolor{c2}{HTML}{FBD9BD}
\definecolor{c3}{HTML}{fe793d}
\definecolor{c4}{HTML}{eedeb0}
\definecolor{rouse}{rgb}{0.981,0.961,0.941}
\usepackage[most]{tcolorbox}
\tcbset{colback=blue!5!white, colframe=blue!20!white, boxrule=0pt, arc=2mm, left=1mm, right=1mm, top=1mm, bottom=1mm}

\definecolor{adptorange}{RGB}{248, 205, 172}
\definecolor{cmpblue}{RGB}{189, 215, 238}
\definecolor{cmpblue}{RGB}{189, 215, 238}

\definecolor{our_red}{RGB}{232,157,160}
\definecolor{our_blue}{RGB}{136,206,230}
\definecolor{our_orange}{RGB}{246,200,168}
\definecolor{our_green}{RGB}{178,211,164}

\definecolor{attn_code0}{RGB}{247,215,200}
\definecolor{attn_code1}{RGB}{238,169,139}
\definecolor{mlp_code0}{RGB}{204,201,221}
\definecolor{mlp_code1}{RGB}{102,95,153}

\definecolor{token_blue}{RGB}{84, 120, 140}

\usepackage{pifont}       
\usepackage{bbding}       
\usepackage{fontawesome}
\usepackage{xspace}

\usepackage{float}

\newlength\savewidth

\newcolumntype{x}[1]{>{\centering\arraybackslash}p{#1pt}}
\newcolumntype{y}[1]{>{\raggedright\arraybackslash}p{#1pt}}
\newcolumntype{z}[1]{>{\raggedleft\arraybackslash}p{#1pt}}

\renewcommand{\paragraph}[1]{\vspace{1mm}\noindent\textbf{#1}}

\usepackage{colortbl}
\usepackage{xcolor}
\usepackage{wrapfig}

\renewcommand{\paragraph}[1]{\vspace{1.25mm}\noindent\textbf{#1}}

\usepackage{listings}

\definecolor{codeblue}{rgb}{0.21, 0.49, 0.74}
\definecolor{codekw}{rgb}{0.35, 0.35, 0.75}
\lstdefinestyle{Pytorch}{
    language = Python,
    backgroundcolor = \color{white},
    basicstyle = \fontsize{9pt}{8pt}\selectfont\ttfamily\bfseries,
    columns = fullflexible,
    aboveskip=1pt,
    belowskip=1pt,
    breaklines = true,
    captionpos = b,
    commentstyle = \color{codeblue},
    keywordstyle = \color{codekw},
}

\definecolor{green}{HTML}{009000}
\definecolor{red}{HTML}{ea4335}

\title{SparkVSR: Interactive Video Super-Resolution via Sparse Keyframe Propagation}
\author[1]{Jiongze Yu}
\author[1]{Xiangbo Gao}
\author[2]{Pooja Verlani}
\author[2]{Akshay Gadde}
\author[2]{Yilin Wang}
\author[2]{Balu Adsumilli}
\author[1]{Zhengzhong Tu}

\affiliation[1]{Texas A\&M University}
\affiliation[2]{YouTube, Google}

\abstract{
Video Super-Resolution (VSR) aims to restore high-quality video frames from low-resolution (LR) estimates, yet most existing VSR approaches behave like black boxes at inference time: users cannot reliably correct unexpected artifacts, but instead can only accept whatever the model produces. 
In this paper, we propose a novel interactive VSR framework dubbed SparkVSR that makes sparse keyframes a simple and expressive control signal. Specifically, users can first super-resolve or optionally a small set of keyframes using any off-the-shelf image super-resolution (ISR) model, then SparkVSR propagates the keyframe priors to the entire video sequence while remaining grounded by the original LR video motion.
Concretely, we introduce a keyframe-conditioned latent-pixel two-stage training pipeline that fuses LR video latents with sparsely encoded HR keyframe latents to learn robust cross-space propagation and refine perceptual details. At inference time, SparkVSR supports flexible keyframe selection (manual specification, codec I-frame extraction, or random sampling) and a reference-free guidance mechanism that continuously balances keyframe adherence and blind restoration, ensuring robust performance even when reference keyframes are absent or imperfect. Experiments on multiple VSR benchmarks demonstrate improved temporal consistency and strong restoration quality, surpassing baselines by up to 24.6\%, 21.8\%, and 5.6\% on CLIP-IQA, DOVER, and MUSIQ, respectively, enabling controllable, keyframe-driven video super-resolution.
Moreover, we demonstrate that SparkVSR is a generic interactive, keyframe-conditioned video processing framework as it can be applied out of the box to unseen tasks such as old-film restoration and video style transfer.
}

\project{\url{https://sparkvsr.github.io/}}
\date{\today}
\contact{Jiongze Yu (\href{mailto:yjz@tamu.edu}{yjz@tamu.edu}), Zhengzhong Tu (\href{mailto:tzz@tamu.edu}{tzz@tamu.edu})}

\begin{document}
\thispagestyle{firstheader}
\maketitle
\pagestyle{plain}

\begin{figure}[tb]
  \centering
  \includegraphics[width = \textwidth]{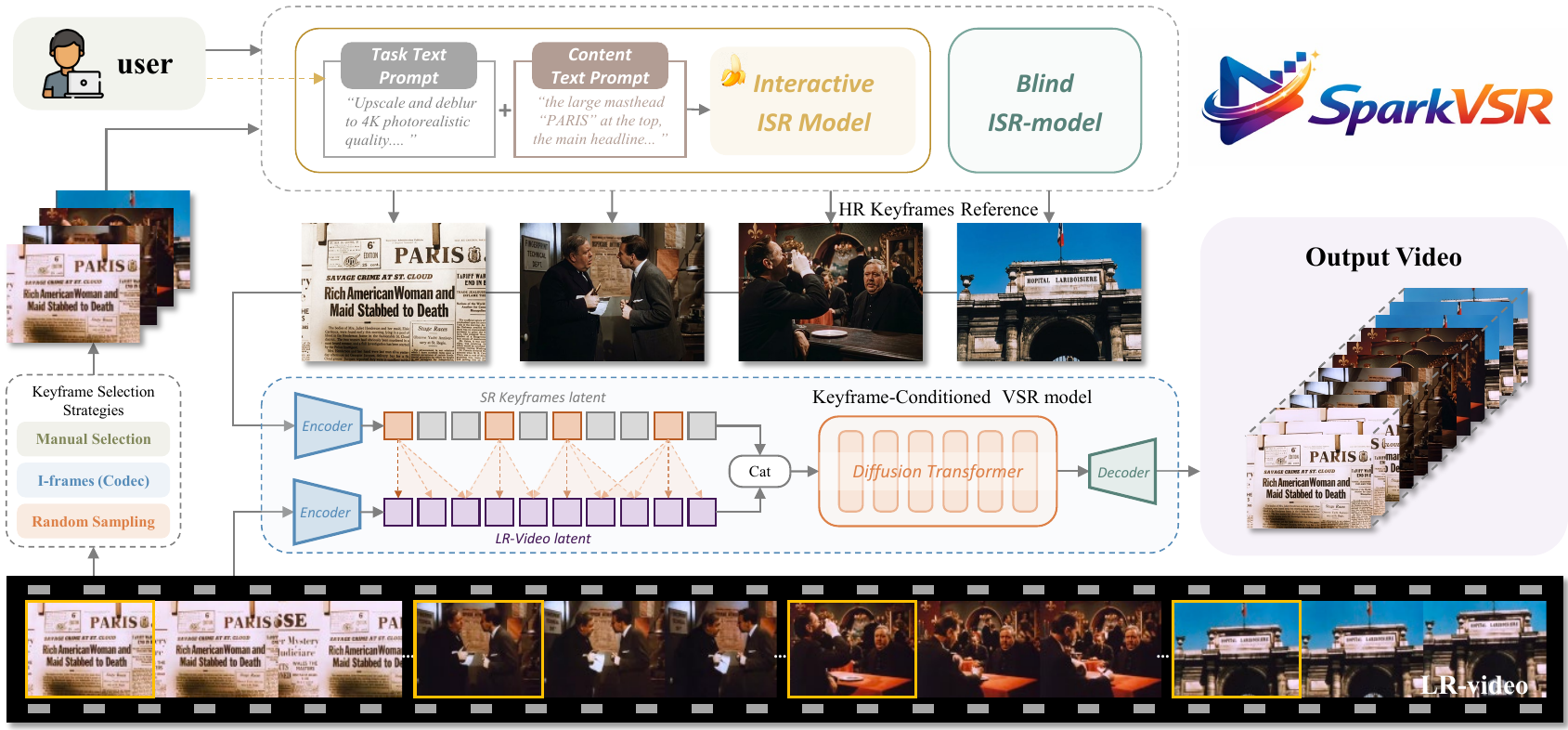}
  \caption{\textbf{Overall inference framework of SparkVSR.} The pipeline consists of three main stages: (1) Keyframe Selection: LR keyframes are extracted using manual, I-frame, or random sampling strategies; (2) HR Reference Generation: Selected frames are upscaled into HR reference keyframes via an interactive (task/content prompt-guided) or blind ISR model; (3) Conditional Video Reconstruction: A Diffusion Transformer-based VSR model fuses the HR keyframe and LR video latents to guide the generation of the final HR video.}
  \label{fig:inference_pipeline}
\end{figure}

\input{sec/1_intro}
\input{sec/2_related_work}
\input{sec/3_method}
\input{sec/4_experiment}
\input{sec/5_conclusion}

{
\small
\bibliographystyle{IEEEtran}
\bibliography{main}
}


\end{document}

%% file: preamble.tex
\definecolor{lightblue}{rgb}{0.0, 0.6, 1.0}
\definecolor{darkgreen}{rgb}{0.0, 0.6, 0.0}
\definecolor{lightgreen}{rgb}{0.1, 0.8, 0.1}
\definecolor{lightred}{rgb}{0.7, 0.7, 0.7}
\definecolor{lightgreen}{rgb}{0, 0, 0}
\definecolor{lightlightgray}{rgb}{0.8, 0.8, 0.8}

\usepackage{titlesec}
\titlespacing*{\paragraph}
{0pt}{0em}{0.2em}  

\makeatletter
\renewcommand\paragraph{\@startsection{paragraph}{4}{\z@}%
  {3pt}
  {-0.5em}
  {\normalfont\normalsize\bfseries}} 
\makeatother

\usepackage{amsmath}  
\usepackage{upgreek}
\usepackage{array}    
\usepackage{multirow}
\usepackage{array}
\usepackage{algorithm}
\usepackage{makecell}
\usepackage{graphicx}
\usepackage{algorithm}
\usepackage{algpseudocode}
\usepackage{colortbl}
\usepackage{enumitem}
\usepackage{wrapfig}
\usepackage{mathrsfs}
\usepackage{pifont}
\usepackage{soul}
\usepackage{amssymb}
\usepackage{tcolorbox}
\usepackage{dirtytalk}
\usepackage{xspace}
\usepackage[utf8]{inputenc}

\usepackage{booktabs}

\usepackage{url}
\definecolor{blue}{rgb}{0.21,0.49,0.74}
\definecolor{red}{rgb}{0.8, 0.2, 0.2}
\definecolor{green}{rgb}{0, 0.5, 0}
\definecolor{yellow}{RGB}{218, 160, 109}
\definecolor{gray}{RGB}{155, 155, 155}

\crefname{section}{Sec.}{Secs.}
\Crefname{section}{Section}{Sections}
\Crefname{table}{Table}{Tables}
\crefname{table}{Tab.}{Tabs.}
\crefname{figure}{Fig.}{Figs.}
\Crefname{figure}{Figure}{Figures}
\crefname{appendix}{App.}{Apps.}
\Crefname{appendix}{Appendix}{Appendices}

%% file: sec/1_intro.tex
\section{Introduction}
\label{sec:intro}
Video Super-Resolution (VSR) workflow plays a critical role in modern computational photography and video production, aiming to restore visually pleasing, high-frequency details from degraded, low-resolution (LR) video sequences \cite{jo2018deep, nah2019ntire}. Despite rapid progress in learning-based VSR models and impressive gains in benchmark metrics~\cite{chen2025dove, wang2025seedvr, wang2025seedvr2, zhuang2025flashvsr, tu2021ugc, tu2021rapique}, the majority of these approaches still operate as black boxes. That is, once trained, users have little ability to steer the inference results, while the model fully determines the outputs. Recent attempts \cite{xie2025star} to introduce controllability via text prompts describing the video content provide only coarse, high-level guidance and often fall short when users need precise, frame-level controls. 
In practice, this ``hope-for-the-best'' inference paradigm (i.e., most existing blind VSR) limits the usability of VSR models in real restoration and content creation workflows where subjective preference and targeted corrections are indispensable.

A central reason why controllability is difficult--- but necessary---is that super-resolution is inherently ill-posed~\cite{jo2021tackling}.
The same LR input can correspond to multiple plausible HR reconstructions that differ in texture, sharpness, and fine appearance, and choosing among these plausible outputs is more of a user-intention problem than a learning problem. This motivates a control interface that is both expressive and lightweight.
In video workflows, we deem keyframes an effective interface: editing or validating a small number of anchor frames is far more practical than supervising every frame, yet those anchors can strongly constrain the overall result if the model can propagate them reliably.

In parallel, single-image super-resolution~\cite{wang2024exploiting, lin2024diffbir, wu2024one, sun2025pixel, qi2024spire, zhang2024prior} (or generic editing~\cite{zuo2025nano}) has recently made dramatic advances, particularly with strong priors to synthesize photorealistic textures and details.
However, applying ISR independently to each frame typically causes severe temporal inconsistency and flicker, because per-frame generation ignores cross-frame dynamics.
More broadly, we observed that VSR models are often lagging behind the best frame-level ISR in per-frame visual quality due to the fact that VSR is forced to learn both (i) spatial priors and (ii) complex temporal consistency, simultaneously.
This further suggests a principled decomposition that leverages a state-of-the-art (possibly interactive) ISR model to obtain high-quality keyframe anchors, and trains a dedicated VSR model to propagate these anchor priors across time while staying faithful to the LR motion structure.

Based on these observations, we propose \textbf{SparkVSR}, an interactive framework for VSR via sparse keyframe propagation.
SparkVSR transforms VSR into a human-in-the-loop process: users (or an automatic policy) select a small set of keyframes, generate high-quality HR keyframe references using any off-the-shelf ISR model, and then SparkVSR propagates these reference priors to reconstruct a temporally consistent HR video, as shown in~\ref{fig:inference_pipeline}.
Specifically, SparkVSR introduces a keyframe-conditioned latent–pixel two-stage training pipeline~\ref{fig:training_pipeline}. We explicitly retain the LR video information by encoding it into latents and concatenating it with sparsely encoded HR keyframe latents, enabling robust cross-space propagation while grounding reconstruction in the input video structure. At inference time, SparkVSR supports flexible keyframe selection (manual, codec I-frame, etc.).
We adopt classifier-free guidance (CFG)~\cite{ho2022classifier} during training to allow the model to adjust the condition strength when reference frames are absent or noisy. This design allows SparkVSR to both improve controllability and stand on top of modern ISR advances, yielding stronger per-frame quality together with temporal consistency.
Extensive evaluations across multiple benchmarks validate these advantages, where SparkVSR outperforms state-of-the-art baselines by up to 24.6\% in CLIP-IQA, 21.8\% in DOVER, and 5.6\% in MUSIQ.
We also demonstrate that our SparkVSR framework exhibits strong task generalization abilities as it can be directly applied to old-film restoration and video style transfer.
Our contributions are summarized as follows:
\begin{itemize}
    \item 
    \textbf{A Novel Interactive VSR Paradigm.} We formulate VSR as an interactive reconstruction process and propose SparkVSR, where sparse, editable keyframes act as controllable anchors, enabling fine-grained correction and customization beyond black-box inference.
    
    \item
    \textbf{Robust Keyframe-Conditioned Latent-Pixel Training.} We propose a two-stage training strategy that fuses LR video latents with sparse HR keyframe latents and refines results in pixel space, equipping the model with robust propagation while maintaining structural fidelity.
    
    \item
    \textbf{Flexible inference with controllable guidance.} We provide practical keyframe selection strategies and a method to trade off keyframe adherence and blind restoration, ensuring robustness across diverse scenarios.

    \item 
    \textbf{State-of-the-art Performance.} We demonstrate that SparkVSR achieves state-of-the-art performance in terms of both full-reference and no-reference metrics (using different modes), and reaches Pareto optimality in the perception-distortion tradeoff diagram.
\end{itemize}

%% file: sec/2_related_work.tex
\section{Related Work}

\subsection{Video Super-Resolution}
The evolution of Video Super-Resolution (VSR) has been fundamentally driven by architectures designed to capture complex spatio-temporal correlations. Early methods relied on implicit temporal aggregation \cite{jo2018deep, nah2019ntire, isobe2020video, li2020mucan}, while subsequent frameworks introduced explicit alignment mechanisms, such as optical flow and deformable convolutions \cite{wang2019edvr, tian2020tdan, chan2021basicvsr, chan2022basicvsr++}, to improve structural reconstruction. More recently, Transformer-based \cite{liang2024vrt, liu2022video, liang2022recurrent} and Diffusion-based VSR models \cite{ho2022video, yang2023diffusion, xie2025star, chen2025dove, wang2025seedvr, wang2025seedvr2, zhuang2025flashvsr} have achieved state-of-the-art visual quality by synthesizing highly realistic textures from complex degradations. However, despite these impressive quantitative gains, contemporary models predominantly operate as deterministic, end-to-end mapping functions. They essentially function as black boxes during inference, lacking the fine-grained, interactive mechanisms necessary for users to actively guide the reconstruction process, correct specific artifacts, or inject customized visual intentions.

\subsection{Controllable Image Super-Resolution}
To overcome the limitations of deterministic restoration, recent Image Super-Resolution (ISR) approaches heavily leverage the generative priors of large-scale diffusion models \cite{saharia2022palette, rombach2022high, ramesh2022hierarchical, wang2024exploiting, lin2024diffbir, wu2024one, sun2025pixel, qi2024spire, zuo2025kagent}. These robust spatial priors enable the synthesis of high-fidelity details from severely degraded inputs—a feat VSR struggles to match due to complex cross-frame dynamics and motion blur \cite{zhang2021learning}. Crucially, modern generative paradigms have introduced unprecedented user controllability. Interactive models, such as Nano-Banana-Pro \cite{google2025nanobananapro, zuo2025nano}, along with text or spatially-guided frameworks \cite{brooks2023instructpix2pix, zhang2023adding, mou2024t2i}, allow users to actively shape the restored output. While these high-quality, customized single frames serve as ideal reference anchors for reconstruction, applying such ISR models independently across video sequences inevitably disrupts the underlying motion dynamics, resulting in severe temporal flickering and structural inconsistency \cite{lai2018learning}.

\subsection{Keyframe-Conditioned Video Processing}
In the broader domain of video generation and editing, utilizing sparse keyframes to guide temporal synthesis has emerged as a highly effective paradigm \cite{wu2023tune, geyer2023tokenflow, qi2023fatezero, liu2025generative, huang2026ffp, liu2025dreamontage, gao2026pisco, wu2026consid, gao2026visualchronometer}. These methods demonstrate that powerful visual priors from individual anchor frames can be robustly propagated across the temporal dimension, often by distributing globally selected keyframes to guide specific local sequence chunks \cite{blattmann2023align}. However, directly adopting these techniques for VSR presents a critical challenge: VSR demands absolute structural fidelity. Existing generative methods often hallucinate content that deviates from original motion constraints, causing severe distortions in the strict LR-to-HR mapping required for restoration \cite{blattmann2023stable, zhou2022magicvideo}. To bridge this gap, SparkVSR introduces a novel keyframe-conditioned latent-pixel training strategy. By seamlessly integrating the high-quality priors of modern ISR models (via sparse keyframes) with the retained LR latents, our framework explicitly equips the model with robust temporal propagation capabilities while strictly anchoring the output to the video's original structural dynamics.

%% file: sec/3_method.tex
\section{Methodology}

This section details the proposed SparkVSR framework, beginning with its overall architecture (Sec.~\ref{sec:architecture}). We then introduce the keyframe-conditioned latent-pixel training for robust prior propagation (Sec.~\ref{sec:training}), and conclude with the interactive inference phase featuring customizable keyframe selection and flexible reference guidance (Sec.~\ref{sec:inference}).

\begin{figure}[tb]
  \centering
  \includegraphics[width = \textwidth]{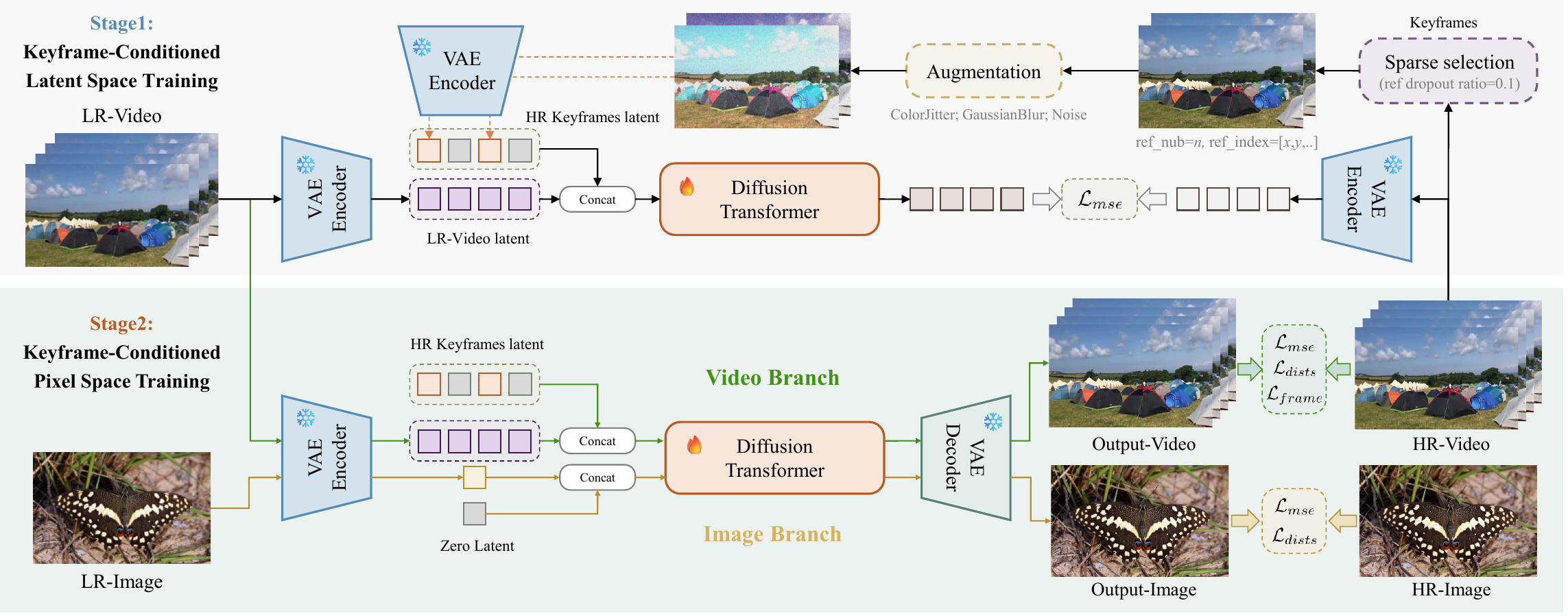}
  \caption{\textbf{Keyframe-conditioned two-stage training pipeline of SparkVSR.} (1) Stage 1 (Latent Space Training): Augmented HR keyframe latents are concatenated with LR video latents to optimize the Diffusion Transformer using $\mathcal{L}_{mse}$. (2) Stage 2 (Pixel Space Training): A joint video-image training mechanism is employed. The video branch is conditioned on HR keyframe latents, while the image branch uses a zero latent. Finally, outputs are decoded by the VAE and refined in the pixel space using mixed losses.}
  \label{fig:training_pipeline}
  \vspace{-10pt}
\end{figure}

\subsection{Overall Architecture}
\label{sec:architecture}
The core of SparkVSR lies in breaking the deterministic black-box mapping of traditional Video Super-Resolution (VSR) by introducing high-quality external reference frames to explicitly guide video generation. To achieve this, we build upon the pretrained weights of the CogVideoX1.5-5B Image-to-Video (I2V) model and design a dual-encoding mechanism to process continuous video sequences and sparse keyframes independently.

\vspace{1mm}
\noindent\textbf{Dual-Encoding and Sparse Reference Generation:}
As illustrated in the inference pipeline (Fig.~\ref{fig:inference_pipeline}), given a degraded low-resolution (LR) video sequence $x_{lr} \in \mathbb{R}^{T \times H \times W \times 3}$, we first encode it into the latent space using the 3D causal VAE inherited from the pretrained model. Due to the temporal downsampling rate of 4 inherent to the 3D VAE, we obtain a 16-channel video latent representation, denoted as $Z_{LR} \in \mathbb{R}^{\frac{T}{4} \times 16 \times H' \times W'}$.

Simultaneously, we generate high-resolution (HR) reference images for the selected sparse keyframes. For interactive restoration, we employ the powerful Nano-Banana-Pro~\cite{google2025nanobananapro, zuo2025nano}, which offers exceptional single-frame generative capabilities. For blind ISR without user intervention, we adopt the current state-of-the-art model, PiSA-SR~\cite{sun2025pixel}. Once the HR keyframes, denoted as $X_{ref}$, are obtained, we design a specific sparse keyframes encoding branch to map these images into the latent space. Crucially, the HR keyframes are sparsely encoded into their corresponding latent indices based on their temporal positions in the video. Let $\mathcal{K}$ denote the set of latent indices corresponding to the selected keyframes. For each latent frame index $i \in \{1, 2, \dots, \frac{T}{4}\}$, the reference latent $Z_{ref}^{(i)}$ is mathematically formulated as:
\begin{equation}
Z_{ref}^{(i)} = 
\begin{cases} 
\mathcal{E}_{sparse}(X_{ref}^{(i)}) & \text{if } i \in \mathcal{K}, \\ 
\mathbf{0} & \text{otherwise}, 
\end{cases}
\end{equation}
where $\mathcal{E}_{sparse}$ denotes the spatial encoder and $\mathbf{0}$ represents a zero tensor of identical spatial and channel dimensions. Consequently, we construct a 16-channel sparse reference latent representation, $Z_{ref} \in \mathbb{R}^{\frac{T}{4} \times 16 \times H' \times W'}$.

\vspace{1mm}
\noindent\textbf{Feature Fusion and One-Step Denoising:}
After obtaining the dual-space features, we concatenate the 16-channel LR video latent $Z_{LR}$ and the 16-channel reference latent $Z_{ref}$ along the channel dimension to form the joint conditional input: $Z_{in} = \text{Concat}(Z_{LR}, Z_{ref}) \in \mathbb{R}^{\frac{T}{4} \times 32 \times H' \times W'}$. 

Following recent diffusion-based VSR paradigms, we initialize the denoising process directly with the encoded LR video latent $Z_{LR}$, treating it as the noisy latent $Z_t$. To minimize computational overhead, we adopt a one-step diffusion strategy inspired by DOVE~\cite{chen2025dove}, specifically setting the timestep to $t=399$. This intermediate step strikes an optimal balance: it preserves sufficient global structure from the LR video while allowing the Diffusion Transformer $v_{\theta}$ to focus exclusively on hallucinating high-frequency details conditioned on $Z_{in}$. Finally, the denoised latent $Z_{sr}$ is decoded by the VAE decoder $\mathcal{D}$ to reconstruct the HR video $x_{sr} = \mathcal{D}(Z_{sr})$.

\subsection{Keyframe-Conditioned Latent-Pixel Training}
\label{sec:training}
To achieve precise keyframe control and effectively propagate the high-quality spatial priors generated by the external ISR model throughout the entire video sequence, we formulate a two-stage Keyframe-Conditioned Latent-Pixel Training strategy, as depicted in the training pipeline (Fig.~\ref{fig:training_pipeline}), inspired by the highly efficient two-stage paradigm~\cite{chen2025dove}.

\vspace{1mm}
\noindent\textbf{Stage 1 (Latent-Space):} 
In the first stage, we fix the VAE decoder and train the Transformer $v_{\theta}$ entirely in the latent space, which significantly improves training efficiency. During the extraction of HR keyframes from the ground-truth video $x_{hr}$, we implement a sparse random selection strategy. Specifically, the total number of selected keyframes is randomly determined with an upper bound of $T/4$, and their temporal indices are randomly sampled with the strict constraint that the interval between any two selected keyframes must be greater than the inherent temporal downsampling rate ($>4$). We then apply severe augmentations (e.g., Color Jitter, Gaussian Blur, Noise) to these selected HR frames to simulate the distribution of external ISR outputs. Guided by the 32-channel concatenated condition $Z_{in}$, the model learns to align and absorb reference features. The optimization objective minimizes the Mean Squared Error (MSE) between the predicted latent $Z_{sr}$ and the HR ground-truth latent $Z_{hr}$.
To preserve robust performance when reference frames are unavailable, we introduce a reference-dropout strategy. During training, the reference latent $Z_{ref}$ is omitted and replaced with zero tensors with a predefined probability $p_{drop}$, compelling the model to perform reference-free blind restoration.

\vspace{1mm}
\noindent\textbf{Stage 2 (Pixel-Space):} 
While latent space training effectively captures the semantic layout, pixel-level constraints are necessary to eliminate temporal flickering and refine perceptual textures. Therefore, we transition to the pixel space and introduce a joint image-video training scheme.

For the \textbf{Video Branch}, the sparse keyframe selection and encoding strategies remain strictly identical to those utilized in Stage 1. The sequence is processed by the network and decoded into the pixel space to generate the output video $\hat{x}_{sr}$. To enforce temporal coherence and exceptional perceptual quality, this branch is supervised by a combination of pixel-wise MSE ($\mathcal{L}_{mse}$), perceptual DISTS loss ($\mathcal{L}_{dists}$), and a frame consistency loss ($\mathcal{L}_{frame}$):
\begin{equation}
\mathcal{L}_{s2-video} = \mathcal{L}_{mse}(\hat{x}_{sr}, x_{hr}) + \lambda_{1}\mathcal{L}_{dists}(\hat{x}_{sr}, x_{hr}) + \lambda_{2}\mathcal{L}_{frame}(\hat{x}_{sr}, x_{hr})
\end{equation}

Simultaneously, for the \textbf{Image Branch}, we process single LR images and explicitly concatenate the encoded image latent with a Zero Latent. This concatenation of the Zero Latent is purposefully designed not only to align the channel dimensions (maintaining the 32-channel input format required by the DiT) but also to substantially solidify the model's robust generative priors in scenarios where reference frames are entirely absent. This branch is optimized using only $\mathcal{L}_{mse}$ and $\mathcal{L}_{dists}$.

\subsection{Flexible Interactive Inference}
\label{sec:inference}
During inference (Fig.~\ref{fig:inference_pipeline}), SparkVSR introduces a highly customizable interactive paradigm that returns control to the user. This flexibility is achieved through versatile keyframe selection strategies, the integration of prompt-guided Image Super-Resolution (ISR) models, and a tunable reference-free guidance mechanism that modulates the influence of spatial priors.

\vspace{1mm}
\noindent\textbf{Flexible Keyframe Selection:}
SparKVSR supports three distinct strategies to accommodate diverse application scenarios:
\begin{enumerate}
    \item \textbf{Manual Selection:} Users can pinpoint specific frames based on aesthetic intentions or target those suffering from the most severe degradation.
    \item \textbf{Codec I-frames:} SparKVSR natively extracts intra-coded I-frames directly from the video stream. As these frames retain maximal spatial information with minimal compression artifacts, they serve as optimal, high-quality anchors for global restoration.
    \item \textbf{Random Sampling:} Designed for automated, large-scale batch processing where human intervention is not required.
\end{enumerate}

\vspace{1mm}
\noindent\textbf{Prompt-Guided Interactive Restoration:}
When employing an interactive ISR model (e.g., Nano-Banana-Pro~\cite{zuo2025nano}), users can exert fine-grained control over keyframe restoration via decoupled textual conditioning. As illustrated in Fig.~\ref{fig:inference_pipeline}, the prompt is systematically divided into two components: a \textbf{Task Text Prompt} (e.g., \textit{``Upscale and deblur to 4K photorealistic quality''}) to specify the core restoration objective, and a \textbf{Content Text Prompt} (e.g., \textit{``the large masthead 'PARIS' at the top''}) to explicitly describe desired semantic or structural details. This human-in-the-loop dual-prompting ensures the generation of pristine, semantically accurate keyframes, especially when tackling extreme degradations where blind models fail. These customized keyframes are subsequently propagated through the SparkVSR architecture to drive the high-fidelity reconstruction of the entire sequence.

\vspace{1mm}
\noindent\textbf{Reference-Free Guidance:}
To grant precise control over feature propagation and balance external keyframe priors with the model's internal generative capacity, we introduce a Reference-Free Guidance mechanism inspired by Classifier-Free Guidance (CFG)~\cite{ho2022classifier}. Benefiting from the reference-dropout strategy and the Zero Latent conditioning, our model inherently supports both referenced and reference-free (blind SR) predictions. During the denoising step, the final prediction $\hat{v}$ is formulated as:
\begin{equation}
\hat{v} = v_{\theta}(Z_{in}^{\text{uncond}}) + s \cdot \left( v_{\theta}(Z_{in}^{\text{cond}}) - v_{\theta}(Z_{in}^{\text{uncond}}) \right)
\end{equation}
where $Z_{in}^{\text{cond}} = \text{Concat}(Z_{LR}, Z_{ref})$ serves as the conditional input, $Z_{in}^{\text{uncond}} = \text{Concat}(Z_{LR}, \mathbf{0})$ replaces reference features with Zero Latents, and $s$ dictates the user-adjustable guidance scale. Specifically, setting $s = 1$ yields standard keyframe-guided generation, whereas $s > 1$ amplifies the high-frequency textures and spatial features from the keyframes, enforcing stronger prior propagation. Conversely, if the external ISR outputs contain slight artifacts, or if the user prefers to rely more heavily on the model's internal blind SR priors, $s$ can be reduced ($s < 1$) or completely disabled ($s = 0$).

%% file: sec/4_experiment.tex
\section{Experiments}

\subsection{Experimental Settings}

\vspace{1mm}
\noindent
\textbf{Datasets.} Following the training data configuration of DOVE~\cite{chen2025dove}, our training corpus combines 2,055 high-resolution video clips from HQ-VSR (degraded via RealBasicVSR~\cite{chan2022investigating}) and 900 images from DIV2K~\cite{cai2019toward} (degraded via Real-ESRGAN~\cite{wang2021real}). For evaluation, we employ synthetic benchmarks (UDM10~\cite{tao2017detail}, SPMCS~\cite{yi2019progressive}, YouHQ40~\cite{zhou2024upscale}) with training-matched degradations, alongside real-world datasets like RealVSR~\cite{yang2021real}, which contains smartphone-captured LQ-HQ pairs. 
Furthermore, we propose \textbf{MovieLQ}, a novel dataset featuring ten 360p ($360\times 480$) vintage film clips from the 1940s to 1950s that cover diverse scenes, including text, human subjects, and landscapes. Each clip lasts 8 seconds at 24 frames per second (fps), totaling 192 frames. Unlike existing benchmarks, it offers longer sequences with complex, authentic degradations inherently originating from historical imaging devices and legacy video compression.

\vspace{1mm}
\noindent
\textbf{Evaluation Metrics.} We comprehensively assess model performance using both image (IQA) and video quality assessment (VQA) metrics. The IQA suite includes standard fidelity metrics (PSNR, SSIM~\cite{wang2004image}) and perceptual indicators (LPIPS~\cite{zhang2018unreasonable}, CLIP-IQA~\cite{wang2023exploring}, MUSIQ~\cite{ke2021musiq}). For VQA, FasterVQA~\cite{wu2023neighbourhood} and DOVER~\cite{wu2023exploring} are utilized to accurately measure the overall spatial-temporal video quality.

\vspace{1mm}
\noindent
\textbf{Implementation Details.} SparkVSR is built upon the CogVideoX1.5-5B I2V~\cite{yang2025cogvideox} foundation model and optimized via a two-stage fine-tuning strategy on four NVIDIA A100-80GB GPUs (total batch size 8) using the AdamW optimizer~\cite{loshchilov2018fixing} ($\beta_1=0.9$, $\beta_2=0.95$, $\beta_3=0.98$). Throughout training, reference frames are sampled with random quantities and temporal intervals (strictly exceeding the VAE's temporal downsampling rate) and augmented via ColorJitter, GaussianBlur, and noise injection. Additionally, we set the reference-dropout probability to $p_{drop} = 0.1$ to robustly maintain zero-reference super-resolution capability. Stage-1 trains exclusively on video sequences ($320\times 640$ resolution, 33 frames) for 10,000 iterations at a learning rate of $2\times 10^{-5}$, while Stage-2 shifts to a joint video-image paradigm ($\varphi=0.5$) for 500 iterations at $5\times 10^{-6}$ with loss weights $\lambda_1=\lambda_2=1$.

\begin{table}[t]
    \centering
    \footnotesize
    \renewcommand{\arraystretch}{1.2}
    \setlength{\tabcolsep}{3pt}
    \caption{\textbf{Quantitative comparison of our method against state-of-the-art methods across multiple datasets.} The \colorbox{bestcolor}{best} and \colorbox{sbestcolor}{second best} results are highlighted. Ours$^*$, Ours$^\dagger$, and Ours$^\ddagger$ denote our method without reference, with Nano-Banana-Pro reference, and with PiSA-SR reference, respectively.}
    \label{tab:main_results_final}
    
    \resizebox{\textwidth}{!}{%
    \begin{tabular}{c | c | C C C C C C | C C C}
        \Xhline{1pt}
        \scriptsize Dataset & \scriptsize Metric & 
        \scriptsize STAR & 
        \scriptsize DOVE & 
        \scriptsize SeedVR2-3B & 
        \scriptsize SeedVR2-7B & 
        \scriptsize FlashVSR-Tiny & 
        \scriptsize FlashVSR-Full & 
        \scriptsize Ours$^*$ & 
        \scriptsize Ours$^\dagger$ & 
        \scriptsize Ours$^\ddagger$ \\
        \hline
        \multirow{7}{*}{UDM10} 
        & PSNR$\uparrow$     & 24.11 & \sbest 26.52 & 25.29 & 25.44 & 23.84 & 23.58 & \best 26.62 & 23.70 & 23.43 \\
        & SSIM$\uparrow$     & 0.7052 & \sbest 0.7697 & 0.7573 & 0.7583 & 0.7095 & 0.6993 & \best 0.7756 & 0.6807 & 0.6710 \\
        & LPIPS$\downarrow$  & 0.3987 & 0.2709 & \sbest 0.2667 & \best 0.2526 & 0.2800 & 0.2915 & 0.2830 & 0.3376 & 0.3548 \\
        & MUSIQ$\uparrow$    & 35.58 & 61.11 & 47.13 & 49.06 & 63.42 & 65.84 & 55.79 & \sbest 66.16 & \best 67.52 \\
        & CLIP-IQA$\uparrow$ & 0.2505 & 0.5011 & 0.2797 & 0.2882 & 0.4587 & 0.5016 & 0.4303 & \sbest 0.5501 & \best 0.6252 \\
        & FasterVQA$\uparrow$& 0.6046 & 0.8155 & 0.5894 & 0.6256 & 0.7707 & 0.7899 & 0.7535 & \best 0.8357 & \sbest 0.8202 \\
        & DOVER$\uparrow$    & 0.3331 & 0.5664 & 0.3716 & 0.4095 & 0.5178 & 0.5317 & 0.5494 & \best 0.6902 & \sbest 0.6411 \\
        \hline
        \multirow{7}{*}{SPMCS} 
        & PSNR$\uparrow$     & 22.71 & \best 23.08 & 22.2 & 22.51 & 21.51 & 21.39 & \sbest 23.04 & 19.68 & 20.12 \\
        & SSIM$\uparrow$     & \best 0.6379 & 0.6190 & 0.6067 & 0.6183 & 0.5463 & 0.5403 & \sbest 0.6237 & 0.4704 & 0.4908 \\
        & LPIPS$\downarrow$  & 0.4558 & \sbest 0.2887 & 0.2888 & \best 0.2782 & 0.2907 & 0.3015 & 0.3139 & 0.3356 & 0.3387 \\
        & MUSIQ$\uparrow$    & 32.32 & 69.11 & 63.73 & 64.06 & 68.2 & 69.48 & 65.53 & \sbest 72.97 & \best 73.38 \\
        & CLIP-IQA$\uparrow$ & 0.2807 & 0.5708 & 0.4288 & 0.4314 & 0.4441 & 0.4738 & 0.4937 & \best 0.6830 & \sbest 0.6811 \\
        & FasterVQA$\uparrow$& 0.5398 & 0.7227 & 0.681 & 0.6911 & 0.7291 & \sbest 0.7306 & 0.6791 & \best 0.7474 & 0.7105 \\
        & DOVER$\uparrow$    & 0.4882 & 0.4858 & 0.4391 & 0.4297 & 0.4856 & 0.4701 & 0.4300 & \best 0.5781 & \sbest 0.5337 \\
        \hline
        \multirow{7}{*}{YouHQ40} 
        & PSNR$\uparrow$     & 22.71 & \best 24.44 & 22.13 & 22.21 & 22.35 & 22.09 & \sbest 24.35 & 22.27 & 21.75 \\
        & SSIM$\uparrow$     & 0.6379 & \best 0.6807 & 0.6338 & 0.6418 & 0.5998 & 0.5912 & \sbest 0.6787 & 0.599 & 0.5786 \\
        & LPIPS$\downarrow$  & 0.4558 & 0.2950 & 0.2975 & \best 0.2802 & \sbest 0.2835 & 0.2894 & 0.3272 & 0.3178 & 0.3501 \\
        & MUSIQ$\uparrow$    & 32.32 & 61.03 & 60.804 & 63.18 & 65.97 & \sbest 68.69 & 55.41 & 64.62 & \best 69.10 \\
        & CLIP-IQA$\uparrow$ & 0.2807 & 0.4867 & 0.4106 & 0.4216 & 0.4644 & 0.5043 & 0.4269 & \sbest 0.5411 & \best 0.6346 \\
        & FasterVQA$\uparrow$& 0.5398 & 0.8477 & 0.8554 & 0.8581 & 0.8547 & \best 0.8655 & 0.8075 & \sbest 0.8592 & 0.8420 \\
        & DOVER$\uparrow$    & 0.4882 & 0.7020 & 0.7041 & 0.7050 & 0.6732 & 0.6783 & 0.6694 & \best 0.7393 & \sbest 0.7315 \\
        \hline
        \multirow{7}{*}{RealVSR} 
        & PSNR$\uparrow$     & 17.30 & \best 22.32 & 21.21 & \sbest 22.13 & 19.84 & 19.62 & 22.00 & 21.35 & 19.72 \\
        & SSIM$\uparrow$     & 0.5253 & \best 0.7301 & 0.7010 & 0.7189 & 0.5613 & 0.5561 & \sbest 0.7222 & 0.6982 & 0.6183 \\
        & LPIPS$\downarrow$  & 0.3268 & 0.1850 & 0.2168 & 0.1998 & 0.2367 & 0.2316 & \sbest 0.1809 & \best 0.1678 & 0.2165 \\
        & MUSIQ$\uparrow$    & 68.61 & 71.69 & 64.71 & 63.92 & 68.02 & 71.12 & 71.84 & \sbest 71.86 & \best 75.44 \\
        & CLIP-IQA$\uparrow$ & 0.3377 & 0.5207 & 0.3013 & 0.2853 & 0.3454 & 0.3927 & 0.5407 & \sbest 0.5575 & \best 0.6216 \\
        & FasterVQA$\uparrow$& 0.7184 & \sbest 0.7929 & 0.7287 & 0.7284 & 0.7514 & 0.7501 & 0.7861 & 0.7727 & \best 0.7946 \\
        & DOVER$\uparrow$    & 0.5672 & 0.6150 & 0.5486 & 0.5241 & 0.5278 & 0.5270 & \sbest 0.6180 & 0.5988 & \best 0.6399 \\
        \hline
        \multirow{4}{*}{MovieLQ} 
        & MUSIQ$\uparrow$    & 59.68 & 60.71 & 49.59 & 45.97 & 64.79 & \sbest 66.38 & 56.34 & 65.48 & \best 68.88 \\
        & CLIP-IQA$\uparrow$ & 0.3731 & 0.5433 & 0.3309 & 0.2894 & 0.5487 & 0.5754 & 0.4622 & \sbest 0.6128 & \best 0.6361 \\
        & FasterVQA$\uparrow$& 0.7611 & 0.7647 & 0.5534 & 0.4184 & 0.78 & 0.7822 & 0.7065 & \sbest 0.797 & \best 0.8028 \\
        & DOVER$\uparrow$    & 0.5696 & 0.5101 & 0.3619 & 0.313 & 0.5485 & 0.5544 & 0.5121 & \sbest 0.6194 & \best 0.6212 \\
        \Xhline{1pt}
    \end{tabular}%
    }
\end{table}

\subsection{Comparisons}

To comprehensively evaluate the effectiveness of our proposed SparkVSR, we compare it against several recent state-of-the-art video super-resolution methods. The evaluated baselines include STAR~\cite{xie2025star}, DOVE~\cite{chen2025dove}, SeedVR2 (3B/7B)~\cite{wang2025seedvr2}, and FlashVSR (Tiny/Full)~\cite{zhuang2025flashvsr}. Regarding our keyframe selection strategy, we select only the initial frame as the reference for short sequences (i.e., UDM10, SPMCS, YouHQ40, and RealVSR), while utilizing codec I-frames as reference anchors for the MovieLQ dataset.

\vspace{1em}
\noindent\textbf{Quantitative Evaluation.} As reported in Table~\ref{tab:main_results_final}, SparkVSR consistently achieves superior performance across five diverse benchmark datasets. Our baseline model without reference (SparkVSR$^{*}$) exhibits strong fidelity, achieving the highest PSNR (26.62) and SSIM (0.7756) on the UDM10 dataset. Furthermore, by incorporating reference priors, our reference-guided variants (SparkVSR$^{\dagger}$ with Nano-Banana-Pro reference and SparkVSR$^{\ddagger}$ with PiSA-SR reference) establish new state-of-the-art results in perceptual and video quality assessment (VQA) metrics. Notably, on the challenging real-world MovieLQ dataset, SparkVSR$^{\ddagger}$ dominates perceptual evaluations, yielding the best scores in MUSIQ (68.88), CLIP-IQA (0.6361), FasterVQA (0.8028), and DOVER (0.6212). This demonstrates the robustness of our reference-guided approach in handling complex real-world degradations.

\begin{figure}[tb]
  \centering
  \includegraphics[width = \textwidth]{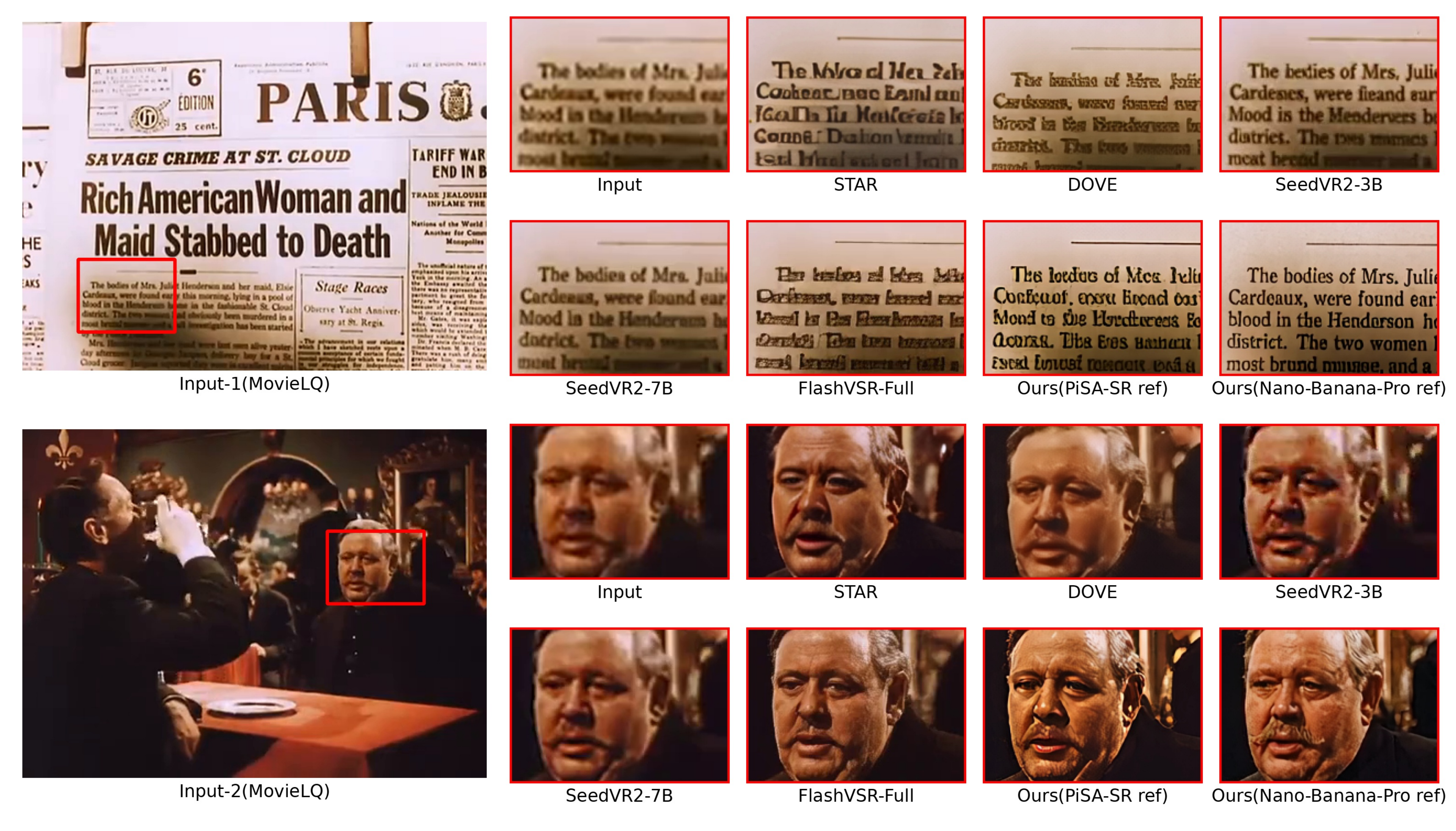}
  \caption{\textbf{Qualitative visual comparisons on the MovieLQ dataset.} Compared to state-of-the-art VSR methods, SparkVSR demonstrates superior recovery of fine textures and structural details, particularly in restoring highly degraded text and facial features.}
  \label{fig:visual_compare_1}
\end{figure}

\begin{figure}[htb]
  \centering
  \includegraphics[width = \textwidth]{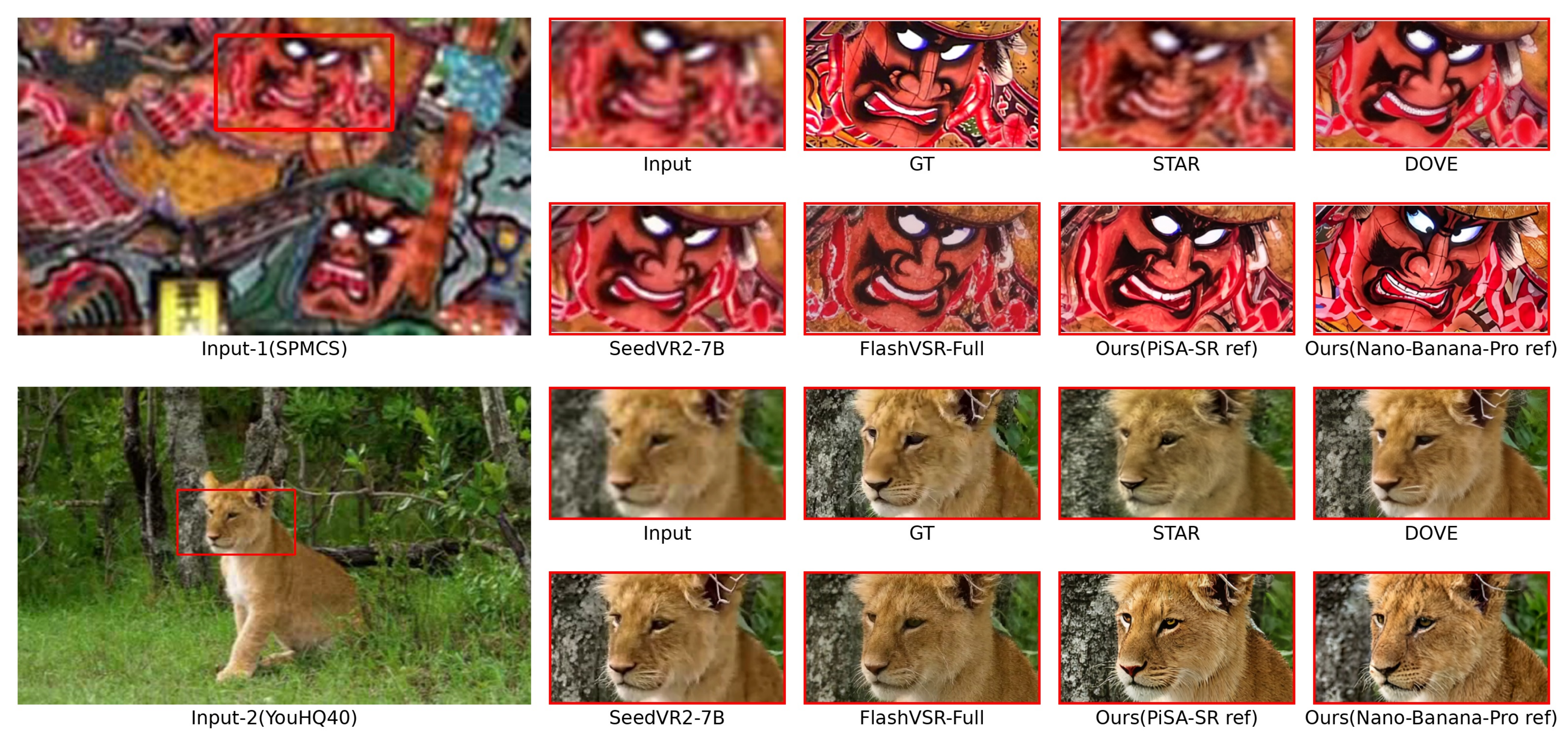}
  \caption{\textbf{Qualitative visual comparisons on the SPMCS~\cite{yi2019progressive} and YouHQ40~\cite{zhou2024upscale} datasets.} We compare our method against recent state-of-the-art VSR models. Guided by high-resolution references, SparkVSR excels in reconstructing sharp edges in animation scenes (top) and fine, realistic textures in natural scenes (bottom).}
  \label{fig:visual_compare_2}
\end{figure}

\vspace{1em}
\noindent\textbf{Qualitative Evaluation.} Compared to existing approaches, SparkVSR consistently produces sharper, more realistic restorations while avoiding undesired artifacts. Specifically, on the real-world MovieLQ dataset (Figure~\ref{fig:visual_compare_1}), our method successfully reconstructs highly legible text and delicate facial details (e.g., skin texture and facial hair), effectively mitigating the severe blurring and over-smoothing prevalent in baselines like DOVE, FlashVSR, and STAR. Furthermore, when evaluated against Ground Truth references on the SPMCS and YouHQ40 datasets (Figure~\ref{fig:visual_compare_2}), SparkVSR exhibits exceptional high-frequency detail regeneration, precisely restoring complex structural edges and fine natural textures (e.g., animal fur).

\subsection{Ablation Study}
To validate the effectiveness of the core components in SparkVSR, we conduct comprehensive ablation studies analyzing our training strategies, perception-distortion control via the Reference-Free Guidance (RFG) scale, and reference frame selection.

\begin{figure}[t]
    \centering
    \captionsetup{font=small}
    \captionsetup[figure]{skip=2pt}
    \captionsetup[table]{skip=2pt}

    \begin{minipage}[t]{0.38\textwidth}
        \vspace{0pt}
        \centering
        {\captionsetup{type=table}\caption*{\phantom{Ablation of training strategies on UDM10 (NBP denotes nano-banana-pro).}}}
        \vspace{-18pt}
        \includegraphics[width=\linewidth]{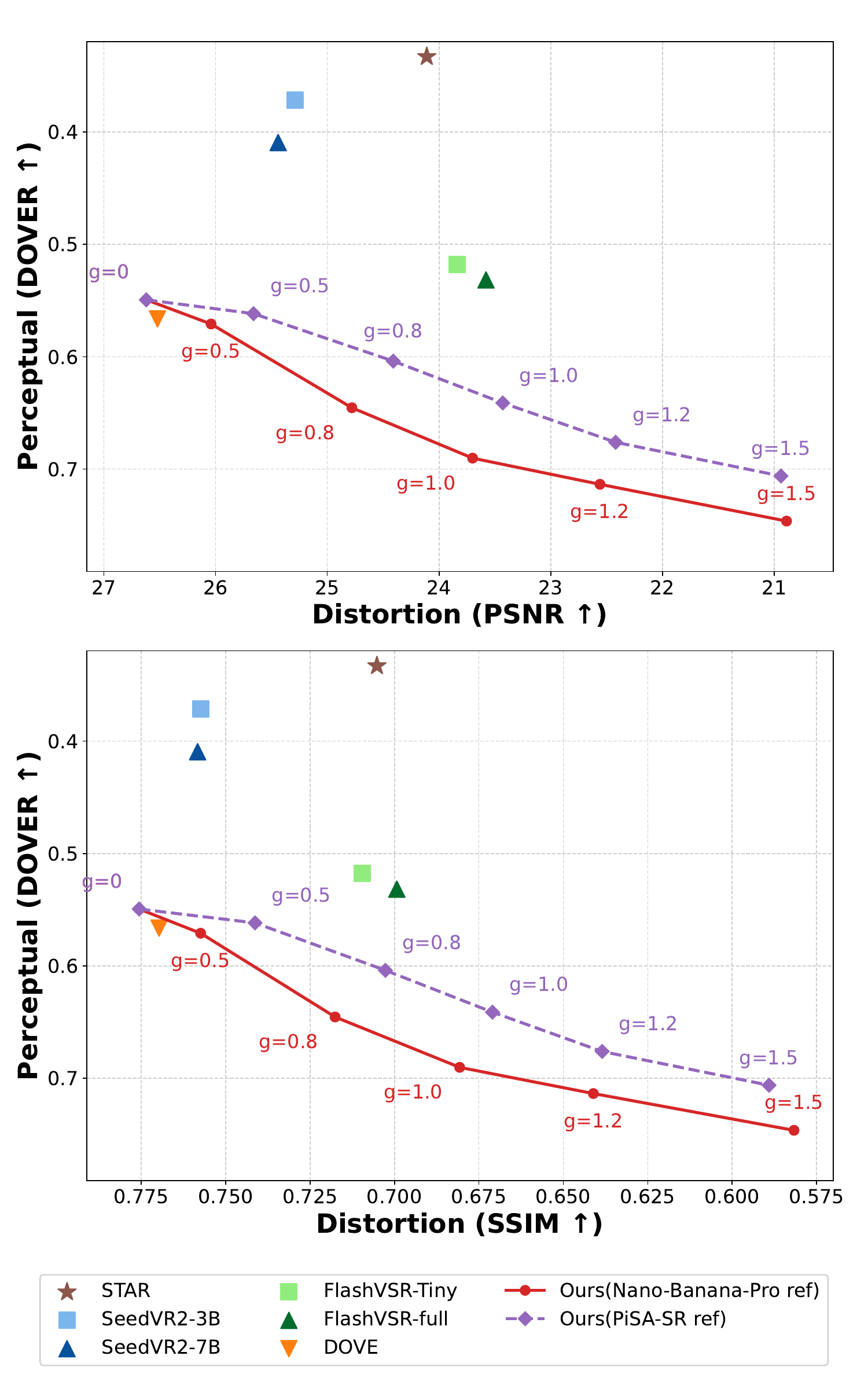}
        \caption{Perception-distortion trade-off. Comparison on PSNR and SSIM vs. DOVER.}
        \label{fig:tradeoff_plot}
    \end{minipage}
    \hfill
    \begin{minipage}[t]{0.58\textwidth}
        \vspace{0pt}
        \centering

        \fontsize{6.7pt}{9.2pt}\selectfont
        \renewcommand{\arraystretch}{1.3}

        \captionof{table}{Ablation of training strategies on the UDM10~\cite{tao2017detail} dataset. "NBP" denotes the Nano-Banana-Pro method. The \colorbox{bestcolor}{best} and \colorbox{sbestcolor}{second best} results are highlighted.}
        \label{tab:ablation_training}
        
        \begin{tabularx}{\linewidth}{l *{6}{Y}}
            \toprule
            & \multicolumn{3}{c}{ours(S1)} & \multicolumn{3}{c}{ours(S1+S2)} \\
            \cmidrule(lr){2-4} \cmidrule(lr){5-7}
            Metric & None & NBP & PiSA & None & NBP & PiSA \\
            \midrule
            PSNR$\uparrow$     & {\cellcolor{bestcolor}26.73} & 24.53 & 24.35 & {\cellcolor{sbestcolor}26.62} & 23.70 & 23.43 \\
            SSIM$\uparrow$     & {\cellcolor{bestcolor}0.778} & 0.707 & 0.698 & {\cellcolor{sbestcolor}0.776} & 0.681 & 0.671 \\
            LPIPS$\downarrow$  & {\cellcolor{sbestcolor}0.330} & 0.338 & 0.354 & {\cellcolor{bestcolor}0.283} & 0.338 & 0.355 \\
            MUSIQ$\uparrow$    & 44.04 & 62.57 & 63.67 & 55.79 & {\cellcolor{sbestcolor}66.16} & {\cellcolor{bestcolor}67.52} \\
            CLIP-IQA$\uparrow$ & 0.331 & 0.474 & 0.537 & 0.430 & {\cellcolor{sbestcolor}0.550} & {\cellcolor{bestcolor}0.625} \\
            FasterVQA$\uparrow$& 0.656 & 0.824 & 0.789 & 0.754 & {\cellcolor{bestcolor}0.836} & {\cellcolor{sbestcolor}0.820} \\
            DOVER$\uparrow$    & 0.439 & {\cellcolor{sbestcolor}0.642} & 0.609 & 0.549 & {\cellcolor{bestcolor}0.690} & 0.641 \\
            \bottomrule
        \end{tabularx}

        \vspace{10pt}

        \captionof{table}{Ablation of the number and positions of reference frames on the MovieLQ dataset. The \colorbox{bestcolor}{best} and \colorbox{sbestcolor}{second best} results are highlighted.}
        \label{tab:ablation_frames}
        
        \begin{tabularx}{\linewidth}{c c *{4}{Y}}
            \toprule
            Ref. Num & Ref. Idx & MUSIQ$\uparrow$ & C-IQA$\uparrow$ & F-VQA$\uparrow$ & DOVER$\uparrow$ \\
            \midrule
            0        & -               & 56.34 & 0.462 & 0.707 & 0.512 \\
            1        & [1]             & 61.73 & 0.535 & 0.764 & 0.562 \\
            2        & [1,192]         & 63.76 & 0.566 & 0.774 & 0.580 \\
            3        & [1,96,192]      & 64.84 & 0.594 & {\cellcolor{sbestcolor}0.795} & 0.596 \\
            4        & [1,64,128,192]  & {\cellcolor{bestcolor}65.76} & {\cellcolor{sbestcolor}0.610} & 0.793 & {\cellcolor{sbestcolor}0.606} \\
            I-frames & [1,48,96,144]   & {\cellcolor{sbestcolor}65.48} & {\cellcolor{bestcolor}0.613} & {\cellcolor{bestcolor}0.797} & {\cellcolor{bestcolor}0.619} \\
            \bottomrule
        \end{tabularx}
    \end{minipage}
\end{figure}

\vspace{1em}
\noindent\textbf{Effectiveness of Training Strategies.} We investigate the impact of our two-stage training paradigm in Table~\ref{tab:ablation_training}. While utilizing only the first stage (S1) achieves high fidelity, evidenced by a PSNR of 26.73 under zero-reference conditions, it yields sub-optimal perceptual quality. Introducing the second stage (S1+S2) significantly improves all perceptual evaluations, regardless of the reference source. This demonstrates that our refinement stage is crucial for enhancing visual realism while maintaining competitive structural integrity.

\begin{figure}[tb]
  \centering
  \includegraphics[width = \textwidth]{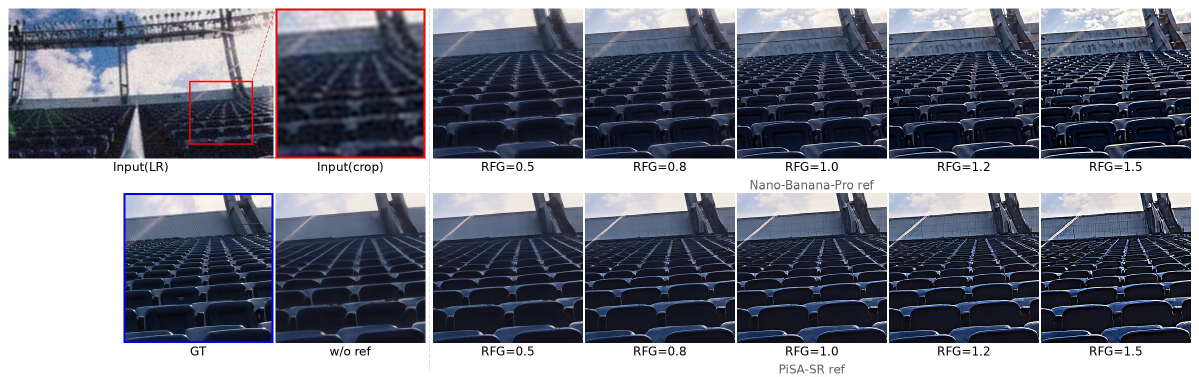}
  \caption{\textbf{Visual ablation study of different reference sources and varying RFG scales.} We compare the restoration results generated by Nano-Banana-Pro and PiSA-SR across a range of Reference-Free Guidance (RFG) values.}
  \label{fig:ablation}
\end{figure}

\vspace{1em}
\noindent\textbf{Perception-Distortion Tradeoff and RFG.} We explore the influence of the Reference-Free Guidance (RFG) scale on restoration results, addressing the inherent mathematical tradeoff between distortion and perceptual quality in image restoration~\cite{blau2018perception}. As shown in Table~\ref{tab:ablation_rfg} and Figure~\ref{fig:tradeoff_plot}, increasing the RFG scale from 0 to 1.5 gradually decreases distortion-based metrics, such as PSNR and SSIM, but substantially boosts perceptual indicators like MUSIQ and CLIP-IQA. Notably, when compared against recent baselines including DOVE, STAR, SeedVR2, and FlashVSR, varying the RFG scale allows SparkVSR to establish a superior Pareto front, as illustrated in Figure~\ref{fig:tradeoff_plot}. Our method consistently achieves better perceptual quality at equivalent distortion levels, effectively pushing the boundaries of this tradeoff. Visually (Figure~\ref{fig:ablation}), although the zero-reference variant successfully recovers fundamental structures with relatively smooth textures, higher RFG scales progressively inject richer high-frequency details, specifically the distinct structural grids of stadium seats and complex natural textures, which closely approximate the ground truth.

\begin{figure}[htb]
  \centering
  \includegraphics[width = \textwidth]{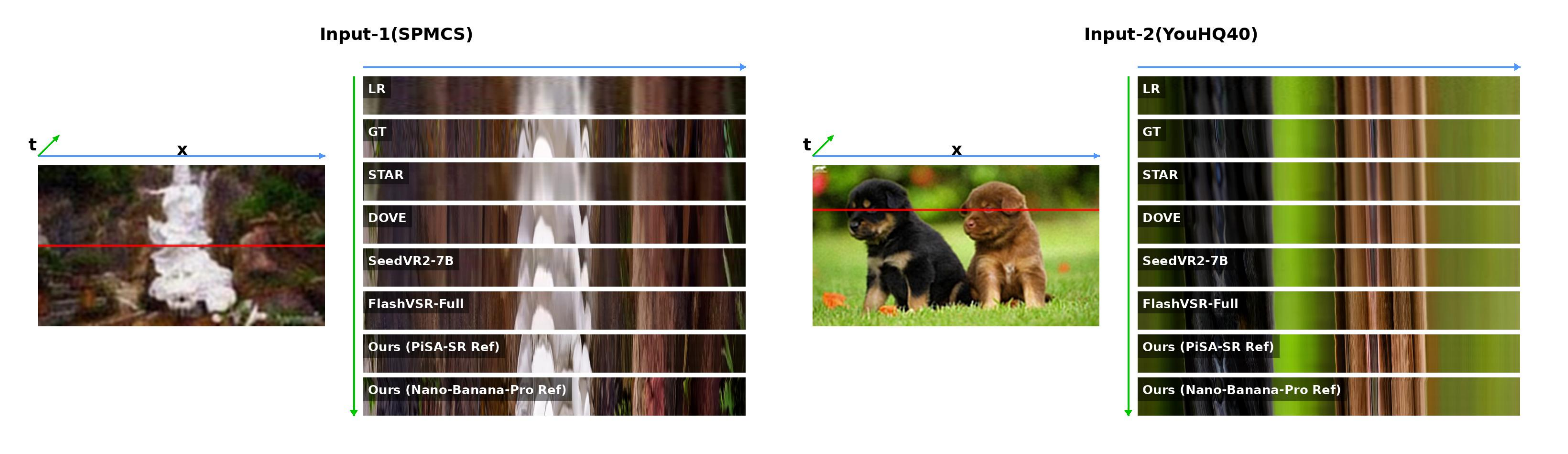}
  \caption{$X-T$ slice profiles comparing different methods on SPMCS (Input-1) and YouHQ40 (Input-2) datasets. The straight and sharp textures in our methods indicate superior temporal stability.}
  \label{fig:xt_slice}
\end{figure}

\begin{table}[tb]
    \centering
    \scriptsize
    \renewcommand{\arraystretch}{1.2} 
    \caption{Ablation study of the impact of various reference sources and varying RFG scales on the UDM10~\cite{tao2017detail} dataset. The \colorbox{bestcolor}{best} and \colorbox{sbestcolor}{second best} results are highlighted.}
    \label{tab:ablation_rfg}
    
    \resizebox{\linewidth}{!}{%
        \begin{tabular}{l r rrrrr rrrrr}
            \toprule
            Ref-source & - & \multicolumn{5}{c}{Nano-Banana-Pro} & \multicolumn{5}{c}{PiSA-SR} \\
            \cmidrule(lr){3-7} \cmidrule(lr){8-12}
            RFG & 0 & 0.5 & 0.8 & 1.0 & 1.2 & 1.5 & 0.5 & 0.8 & 1.0 & 1.2 & 1.5 \\
            \midrule
            PSNR$\uparrow$     & {\cellcolor{bestcolor}26.62} & {\cellcolor{sbestcolor}26.04} & 24.78 & 23.70 & 22.56 & 20.89 & 25.66 & 24.41 & 23.43 & 22.42 & 20.94 \\
            SSIM$\uparrow$     & {\cellcolor{bestcolor}0.776} & {\cellcolor{sbestcolor}0.757} & 0.718 & 0.681 & 0.641 & 0.582 & 0.741 & 0.703 & 0.671 & 0.639 & 0.589 \\
            LPIPS$\downarrow$  & {\cellcolor{bestcolor}0.283} & {\cellcolor{sbestcolor}0.292} & 0.312 & 0.338 & 0.367 & 0.412 & 0.312 & 0.337 & 0.355 & 0.372 & 0.399 \\
            MUSIQ$\uparrow$    & 55.79 & 58.76 & 63.93 & 66.16 & 67.52 & 68.36 & 60.17 & 65.23 & 67.52 & {\cellcolor{sbestcolor}69.01} & {\cellcolor{bestcolor}70.39} \\
            CLIP-IQA$\uparrow$ & 0.430 & 0.447 & 0.515 & 0.550 & 0.570 & 0.576 & 0.496 & 0.584 & 0.625 & {\cellcolor{sbestcolor}0.647} & {\cellcolor{bestcolor}0.652} \\
            FasterVQA$\uparrow$& 0.754 & 0.794 & 0.827 & 0.836 & {\cellcolor{sbestcolor}0.850} & {\cellcolor{bestcolor}0.860} & 0.769 & 0.796 & 0.820 & 0.824 & 0.832 \\
            DOVER$\uparrow$    & 0.549 & 0.571 & 0.645 & 0.690 & {\cellcolor{sbestcolor}0.714} & {\cellcolor{bestcolor}0.746} & 0.562 & 0.604 & 0.641 & 0.676 & 0.706 \\
            \bottomrule
        \end{tabular}%
    } 
\end{table}

\begin{figure}[tb]
  \centering
  \includegraphics[width = \textwidth]{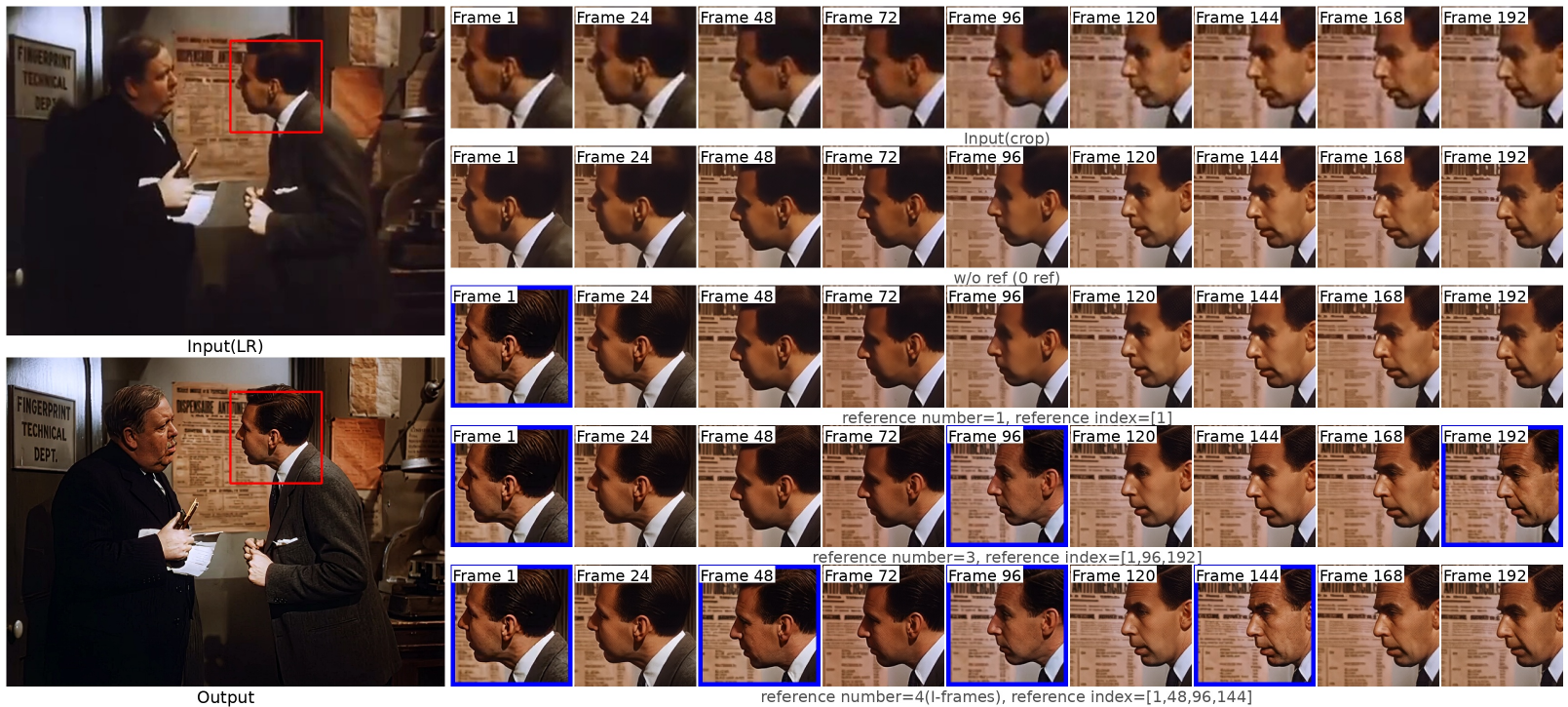}
  \caption{\textbf{Visual ablation study of the number and temporal positions of reference frames.} Blue boxes indicate the specific temporal indices corresponding to the provided high-resolution reference frames, which are generated using Nano-Banana-Pro.}
  \label{fig:ablation_ref_num}
\end{figure}

\vspace{1em}
\noindent\textbf{Influence of Reference Frame Selection.} Table~\ref{tab:ablation_frames} and Figure~\ref{fig:ablation_ref_num} evaluate various reference frame selection strategies. Incorporating a single reference frame significantly boosts perceptual quality (e.g., MUSIQ improves from 56.34 to 61.73). Visually, while the zero-reference baseline yields relatively soft details, injecting a single reference dramatically sharpens the subject, effectively propagating the high-fidelity textures of the keyframe to adjacent frames. Increasing the reference count further enhances performance. Utilizing multiple distributed references (such as three uniformly sampled indices or four extracted I-frames) ensures temporally consistent, delicate textures—specifically hair strands and facial features—across the entire sequence. Achieving optimal and comparable scores across these diverse indices demonstrates that SparkVSR robustly accommodates flexible strategies, encompassing user-defined, random, and codec-aware frame extraction.

\vspace{1em}
\noindent\textbf{X-T Slice Profile Analysis}
To further evaluate temporal consistency, we extract $X-T$ slice profiles by stacking a fixed horizontal scanline across consecutive frames along the temporal axis ($T$), as illustrated in Figure~\ref{fig:xt_slice}. In these profiles, smooth and continuous vertical textures indicate high temporal stability, while jagged or blurry lines reveal flickering and inconsistency. As observed in the profiles for the SPMCS~\cite{yi2019progressive} and YouHQ40~\cite{zhou2024upscale} datasets, the low-resolution (LR) inputs and methods like STAR and DOVE yield continuous but excessively blurry slices, failing to recover sharp structural boundaries. Conversely, while approaches such as SeedVR2-7B and FlashVSR-Full restore finer spatial details, their $X-T$ slices exhibit wavy and jagged edges, indicating inter-frame instability and temporal jitter. In contrast, our proposed methods---Ours (PiSA-SR Ref) and Ours (Nano-Banana-Pro Ref)---produce $X-T$ slices that closely match the Ground Truth (GT). The profiles exhibit sharp edges alongside highly straight and continuous trajectories along the temporal axis, demonstrating that our approach effectively synthesizes high-frequency details while strictly maintaining temporal coherence and suppressing inter-frame flickering.

\subsection{Broader Applications}
The reference-guided architecture of SparkVSR extends beyond standard super-resolution, serving as a robust temporal propagation engine for broader low-level video editing tasks. By utilizing sparsely edited keyframes (e.g., colorized, deblurred, or stylized) as references, SparkVSR effectively propagates high-fidelity features across the entire sequence with rigorous temporal consistency. This unlocks several zero-shot applications without task-specific retraining:

\begin{itemize}
    \item \textbf{Old Video Restoration and Colorization:} Given a few manually restored keyframes from severely degraded historical videos, SparkVSR faithfully propagates clean textures and realistic colors to overcome complex real-world degradations.
    
    \item \textbf{Stylized Video Generation:} Applying artistic edits (e.g., pixel anime art style) to reference frames enables the synthesis of temporally coherent stylized videos while strictly preserving original structural motions.
\end{itemize}

These capabilities demonstrate that SparkVSR transcends a standard super-resolution baseline, emerging as a highly generalizable framework for consistent video feature propagation.

%% file: sec/5_conclusion.tex
\section{Conclusion}
In this paper, we propose SparkVSR, an interactive framework that transforms Video Super-Resolution from a deterministic black box into a controllable, user-guided process. By utilizing sparse, editable keyframes as anchors, SparkVSR efficiently propagates the robust spatial priors of modern image super-resolution models. Our keyframe-conditioned latent-pixel training ensures high-fidelity temporal consistency across the sequence while strictly preserving original motion dynamics. Coupled with flexible keyframe selection and a tunable reference-free guidance mechanism, SparkVSR empowers users to precisely balance perceptual quality and structural fidelity. Extensive experiments demonstrate that SparkVSR achieves state-of-the-art performance across multiple benchmarks and generalizes seamlessly to zero-shot applications like old-film restoration and stylized video generation, establishing a highly versatile paradigm for video reconstruction.